\documentclass{article}

     \PassOptionsToPackage{round}{natbib}

\usepackage[dblblindworkshop, final]{neurips_2025}

\usepackage[utf8]{inputenc} %
\usepackage[T1]{fontenc}    %
\usepackage[hidelinks]{hyperref}       %
\usepackage{url}            %
\usepackage{booktabs}       %
\usepackage{amsfonts}       %
\usepackage{nicefrac}       %
\usepackage{microtype}      %
\usepackage[dvipsnames]{xcolor}         %

\hypersetup{
    colorlinks=true,
    linkcolor=black,  
    urlcolor=Orchid,
    citecolor=RoyalBlue,
    } %
\usepackage[bbgreekl]{mathbbol}
\DeclareSymbolFontAlphabet{\mathbbm}{bbold}
\DeclareSymbolFontAlphabet{\mathbb}{AMSb}
\usepackage{amsmath}
\usepackage{amssymb}
\usepackage[acronym]{glossaries}
\usepackage[pdftex]{graphicx}
\usepackage{subcaption}
\usepackage[ruled,noend]{algorithm2e}
\usepackage{etoolbox}
\makeatletter
\patchcmd{\@algocf@start}%
  {-1.5em}%
  {0pt}%
  {}{}%
\makeatother
\AtBeginDocument{
  \newcommand{\cref}[1]{%
    \ifglsentryexists{#1}{%
      \hypersetup{linkcolor=black}%
    }{%
      \hypersetup{linkcolor=Plum}\ref{#1}%
    }%
    \hypersetup{linkcolor=black}%
  }
}

\newcommand{\todo}[1]{\textcolor{red}{#1}}

\usepackage{amsmath,amsfonts,bm}

\def\VS{{\mathit{VS}}}

\def\eqref#1{equation~\ref{#1}}

\def\1{\bm{1}}

\def\rvtheta{{\bm{\theta}}}

\def\rvphi{{\bm{\phi}}}

\def\rva{{\mathbf{a}}}

\def\rvg{{\mathbf{g}}}

\def\rvs{{\mathbf{s}}}

\def\rvx{{\mathbf{x}}}

\def\rvmu{{\bm{\mu}}}

\def\rmK{{\mathbf{K}}}
\def\rmL{{\mathbf{L}}}
\def\rmM{{\mathbf{M}}}

\def\rmX{{\mathbf{X}}}

\def\rmSigma{{\mathbf{\Sigma}}}

\DeclareMathAlphabet{\mathsfit}{\encodingdefault}{\sfdefault}{m}{sl}
\SetMathAlphabet{\mathsfit}{bold}{\encodingdefault}{\sfdefault}{bx}{n}

\def\gF{{\mathcal{F}}}

\def\gO{{\mathcal{O}}}

\def\gU{{\mathcal{U}}}

\def\sA{{\mathbb{A}}}

\def\sG{{\mathbb{G}}}
\def\sH{{\mathbb{H}}}

\def\sR{{\mathbb{R}}}
\def\sS{{\mathbb{S}}}

\def\sX{{\mathbb{X}}}

\def\sTheta{\mathbbm{\Theta}}

\newcommand{\E}{\mathbb{E}}

\newacronym{im}{IM}{intrinsic motivation}
\newacronym{rl}{RL}{reinforcement learning}
\newacronym[longplural={Markov decision processes}]{mdp}{MDP}{Markov decision process}
\newacronym[longplural={goal-augmented Markov decision processes}]{gamdp}{GAMDP}{goal-augmented Markov decision process}
\newacronym{pomdp}{POMDP}{partially observable Markov decision process}
\newacronym{mi}{MI}{mutual information}
\newacronym{misl}{MISL}{mutual information skill learning}
\newacronym{bmi}{BMI}{behavioural mutual information}
\newacronym{gcrl}{GCRL}{goal-conditioned reinforcement learning}
\newacronym{ued}{UED}{unsupervised environment design}
\newacronym{wurl}{WURL}{Wasserstein Unsupervised
Reinforcement Learning}
\newacronym{metra}{METRA}{Metric-Aware Abstraction}
\newacronym{csf}{CSF}{Contrastive Successor Features}

\newacronym{mmd}{MMD}{maximum mean discrepancy}

\newacronym{lgr}{LGR}{Latent Goal Reaching}
\newacronym{snd}{SND}{System Neural Diversity}
\newacronym{qd}{QD}{Quality-Diversity}

\newacronym{pbsf}{PBSF}{potential-based shaping function}
\newacronym{lp}{LP}{Learning Progress}

\newacronym{sac}{SAC}{soft actor-critic}

\newacronym{credit}{CRediT}{Contributor Role Taxonomy}

\title{VendiRL: A Framework for Self-Supervised Reinforcement Learning of Diversely Diverse Skills}
\workshoptitle{Scaling Environments for Agents (SEA)}

\author{%
  Erik M.~Lintunen\\
  Department of Computer Science\\
  Aalto University, Finland \\
  \texttt{erik.lintunen@aalto.fi} \\
}

\begin{document}

\maketitle

\begin{abstract}
In self-supervised reinforcement learning (RL), one of the key challenges is learning a diverse set of skills to prepare agents for unknown future tasks.
Despite impressive advances, scalability and evaluation remain prevalent issues.
Regarding scalability, the search for meaningful skills can be obscured by high-dimensional feature spaces, where relevant features may vary across downstream task domains.
For evaluating skill diversity, defining what constitutes ``diversity'' typically requires a hard commitment to a specific notion of what it means for skills to be diverse, potentially leading to inconsistencies in how skill diversity is understood, making results across different approaches hard to compare, and leaving many forms of diversity unexplored.
To address these issues, we adopt a measure of sample diversity that translates ideas from ecology to machine learning---the \emph{Vendi Score}---allowing the user to specify and evaluate any desired form of diversity.
We demonstrate how this metric facilitates skill evaluation and introduce \emph{VendiRL}, a unified framework for learning diversely diverse sets of skills.
Given distinct similarity functions, VendiRL motivates distinct forms of diversity, which could support skill-diversity pretraining in new and richly interactive environments where optimising for various forms of diversity may be desirable.
\end{abstract}

\begin{figure}[ht]
    \centering
    \includegraphics[width=1.00\textwidth]{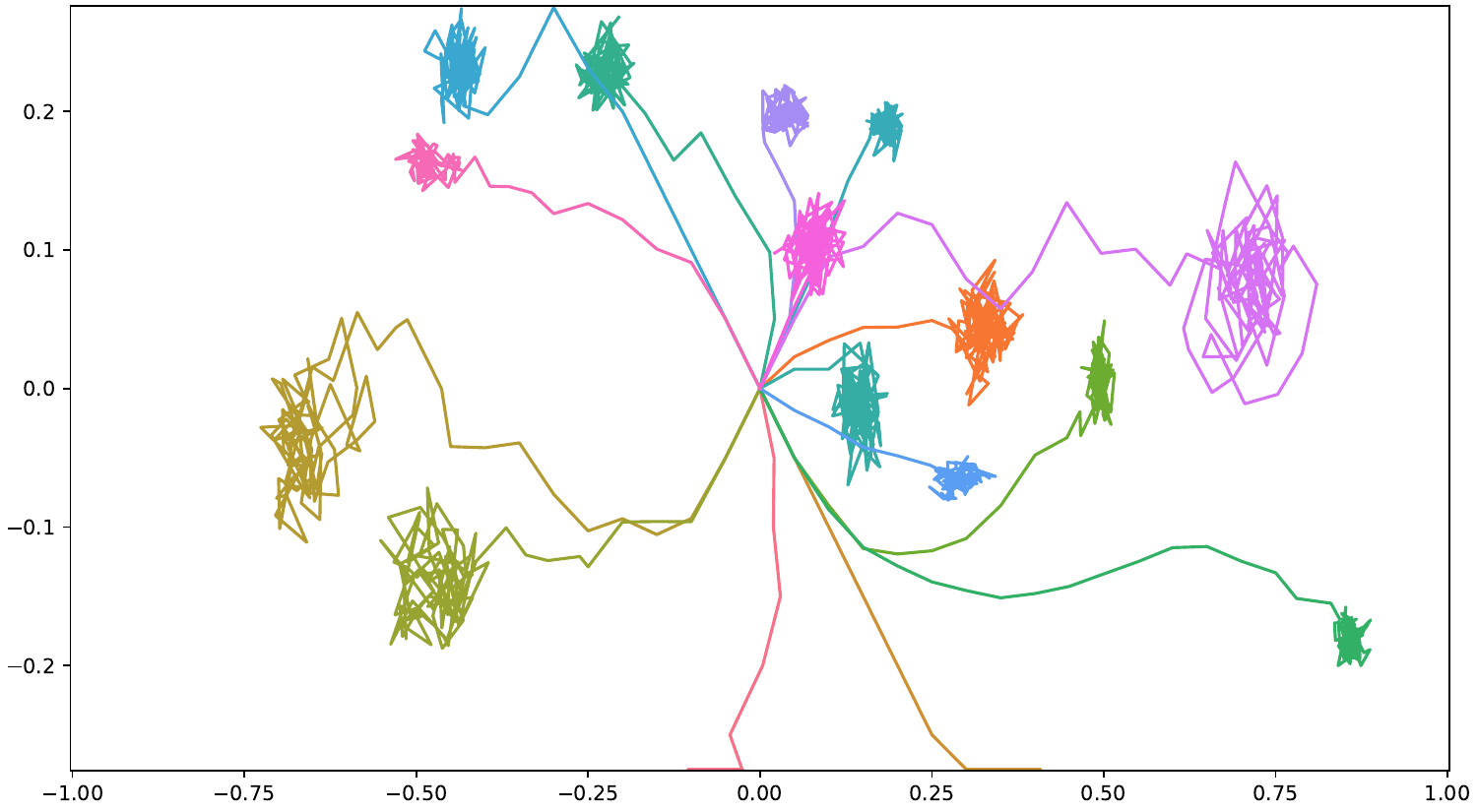}
    \caption{
    A playful example of \emph{VendiRL}-motivated skills in a 2D environment.
    An agent optimised its skills for diversity using a linear combination of two distinct skill-similarity measures, %
    illustrating our framework’s flexibility in targeting diversity---not only individual notions, but also combinations.
    }
    \label{fig:teasercombination}
\end{figure}

\section{Introduction}

In deep \gls{rl}, it is common to use various pretraining tasks to learn transferable knowledge and skills.
Central to learning transferable skills is the challenge of acquiring a \emph{diverse} set of skills.
This area of research has been popularised by impressive results showing that diverse skills can empower agents in adapting to a wide range of future tasks (e.g., \citealp[p.~11]{gregor2016variational}; \citealp[pp.~6--7]{eysenbach2019diversity}; \citealp[pp.~6--7]{hansen2020fast}; \citealp[pp.~8--10]{sharma2020dynamics-aware}; \citealp[p.~4]{laskin2021urlb}; \citealp[p.~6737]{baumli2021relative}; \citealp[pp.~8--9]{shafiullah2022one}; \citealp[pp.~7--8]{yang2023behavior}; \citealp[pp.~8--10]{park2024metra}).
Additionally, some of these successes have been examined theoretically, with recent work suggesting that such approaches can be used to learn an optimal initialisation for unknown reward functions (\citealp{eysenbach2022the}; cf., \citealp{yang2024task}) by facilitating the discovery of the ground-truth features of an environment \citep{reizinger2025skill}.

Despite these advances, it remains unclear how well existing methods scale in complex and richly interactive environments---both virtual and physical---that are characteristic of many applications.
Due to resource constraints, such as training time and compute, these methods often rely on various forms of user supervision.
That is, users apply prior knowledge of downstream tasks to more effectively discover meaningfully diverse skills.
Useful priors include determining the appropriate number of skills to learn, identifying subsets of the feature space known to facilitate learning a set of target behaviours, and choosing a skill-diversity objective that motivates diversity in a desirable way.

We focus on the latter: defining an objective for diversity often requires a hard commitment to a specific understanding of what it means for skills to be ``diverse'', such as those that are discriminable \citep{gregor2016variational}, costly to transform into one another \citep{he2022wasserstein}, or temporally distant \citep{park2024metra}.
Such variance can lead to inconsistencies in how skill diversity is conceptualised and measured---a construct lacking objective evaluation metrics.\footnote{The lack of benchmarks and objective evaluation metrics is not unique to skill diversity; it is a prevalent issue in the broader field of open-ended learning. For a discussion on this topic, see \citet[pp.~1167--1169]{colas2022autotelic}.}
Moreover, these definitions may fail to represent the wide spectrum of possible interpretations of diversity in the real world, many of which could turn out to be beneficial for modelling and evaluating aspects of open-ended learning.

To this end, we introduce \emph{VendiRL}, a framework inspired by the \emph{Vendi Score} \citep{friedman2023the}---a measure of sample diversity that connects ideas from ecology to machine learning.
The Vendi Score enables the specification and evaluation of any desired form of diversity as defined by a similarity function.
Our contributions are three-fold:
(1) we adapt the Vendi Score to measuring skill diversity, resulting in an interpretable and intuitive evaluation metric for diversity in open-ended \gls{rl};
(2) a novel framework for the acquisition of diverse skills through plug-in objectives, specifically via distinct similarity functions and their combinations (as exemplified in Figure~\cref{fig:teasercombination}), enabling the discovery of \emph{diversely} diverse skills under a single reward formalism; and
(3) this ``pick-and-mix'' approach to defining diversity could potentially support the scalability of skill-diversity pretraining to new and richly interactive environments by offering a variety of diversity rewards to choose from.

\section{Preliminaries}

To support the understanding of our work, we introduce the following concepts:
reward functions (Section~\cref{sec:bgtasks}),
skills (Section~\cref{sec:bggcrl}),
intrinsic rewards (Section~\cref{sec:bgimrl}), and
the Vendi Score (Section~\cref{sec:vendiscore}).
Put succinctly, a task for a learning agent is defined by a reward function; skills are conceptualised as a goal-conditioned policy; an agent can self-supervise the process of learning diverse skills by generating its own intrinsic reward signal; and the Vendi Score offers machine learning researchers and practitioners an interpretable, ecologically inspired evaluation metric for any form of diversity.

\subsection{Tasks are defined by reward functions}
\label{sec:bgtasks}

Conceptually, at the core of our work is the idea of tasking an agent to learn a diverse set of skills.
To this end, we first elaborate on our understanding of a ``task'' in the context of \gls{rl}.
Central to \gls{rl} is the \emph{reward function},
mapping from interactions involving states, actions, and successor states to scalar rewards.
This construct is anchored in the \emph{reward hypothesis}---a belief strongly influencing the field---suggesting that ``all of what we mean by goals and purposes can well be thought of as the maximization of the expected value of the cumulative sum of a received scalar signal (called reward)'' (\citealp[p.~53]{sutton2018reinforcement}; cf., \citealp[p.~4]{silver2021reward}).
In this sense, a reward function can be understood as defining a \emph{task} for a learning agent \citep[p.~1]{ng1991policy}.

\subsection{Skills are policies conditioned on goals}
\label{sec:bggcrl}

In \gls{rl}, ``skills'' can be conceptualised as policies trained to accomplish specific tasks.
The challenge of learning a set of multiple skills can be formalised as a \gls{gamdp} within the framework of \acrlong{gcrl} (\acrshort{gcrl}; for a review see \citealp[]{liu2022goal-conditioned}).
This extends the standard definition of a reward function to be conditioned on goals:
\begin{align*}
r : \sS \times \sA \times \sS \times \sG \rightarrow \sR.
\end{align*}
Henceforth, $\rvs, \rvs' \in \sS$ represent a state and successor state, respectively, from the set of possible states, $\rva \in \sA$ represents an action from the set of possible actions, and $\rvg \in \sG$ represents a \emph{goal} from the set of possible goals.
This goal, or \emph{goal-defining variable}, is a parameter to the reward function (cf., \citealp[p.~1165]{colas2022autotelic}; \citealp[p.~6]{aubret2023information}); it indicates which reward function the agent is aiming to maximise.
Then, a \emph{skill}, $\pi(\rva\mid \rvs,\rvg)$, is a policy given a goal, optimising for some notion of cumulative reward according to the goal-conditioned reward.
In this sense, goals can be viewed as ``a set of \emph{constraints} ... that the agent seeks to respect'' \citep[p.~1165, emphasis in original]{colas2022autotelic}.

While the most immediate intuition of a goal is often as a desired state for the agent to reach (e.g., \citealp[p.~1]{kaelbling1993learning}), the formalism allows for a more general set of constraints on behaviour.
In effect, any behaviour that can be defined by attempting to maximise some reward function on the environment can be formulated as a goal--skill pairing.
For a concise typology of goal representations in the intrinsically motivated \acrshort{gcrl} literature, see, for example, \citet[p.~1171--1174]{colas2022autotelic}.

\subsection{Intrinsic rewards enable self-supervised learning}
\label{sec:bgimrl}

Intrinsically motivated \gls{rl} deviates from traditional \gls{rl} by employing \emph{intrinsic} reward functions to generate pseudo-rewards (\citealp[p.~3]{lidayan2025bamdp}).
These intrinsic reward functions evaluate agent-internal variables (\citealp[p.~3]{oudeyer2008how}; cf., \citealp[p.~246]{berlyne1965structure} and \citealp{oudeyer2007what}), which enables their applicability across a diverse range of environments.
Various intrinsic rewards have been explored, such as novelty, learning progress, and empowerment, primarily due to their effectiveness in supporting open-ended development, task-agnostic learning, and the ability to deal with sparse rewards \citep[p.~1161]{colas2022autotelic}.
Models of intrinsic motivation have also been used to explain various aspects of ``wet'' (biological) \gls{rl} (e.g., \citealp{brandle2023empowerment,molinaro2024latent,modirshanechi2025an}), results potentially applicable for the development of truly open-ended and human-like \gls{rl} agents.
For reviews of intrinsically motivated \gls{rl}, see \citet{oudeyer2007what,linke2020adapting,colas2022autotelic,aubret2023information,lidayan2025bamdp}.

A typical feature of intrinsic rewards is their non-stationary nature, which induces a \gls{pomdp} where the dynamics of the reward distribution are unobserved by the agent.
For example, consider rewards computed as a function of the parameters, $\rvphi$, of a learned neural network; then, the reward $r(\rvs,\rva,\rvs',\rvg,\rvphi_t)$ is likely to differ from $r(\rvs,\rva,\rvs',\rvg,\rvphi_{t+1})$.
This characteristic is also prevalent in approaches for learning diverse skills, exemplified in Appendix~\cref{appendix:varmislformal}.

\subsection{A measure of sample diversity inspired by ecology: the Vendi Score}
\label{sec:vendiscore}

A long line of work in ecology concerns a fundamental conceptual problem: how can diversity be quantified in a meaningful way?
Some interpretations emphasise the importance of \emph{abundance} while others the importance of \emph{balance} \citep[pp.~4--5]{leinster2021entropy}.
One choice is to quantify how \emph{different} the species in a community are.
The Vendi Score \citep[]{friedman2023the} bridges this understanding to machine learning, %
representing ``the effective number of dissimilar elements in a sample'' (p.~6).

The Vendi Score is defined as the exponential of the Shannon entropy of the eigenvalues of a kernel (sample-similarity) matrix.
A useful feature of the quantity is that it is interpretable: zero entropy results in the effective number $1$ ($\Leftrightarrow$ all elements are equal) and maximum entropy returns the number of elements $n$ ($\Leftrightarrow$ all elements are effectively unique).
Formally \citep[pp.~5--6]{friedman2023the}:

\textbf{Definition 1 (Vendi Score).} Let $\rvx_1, \cdots, \rvx_n \in \sX$ denote a collection of $n$ samples, let $k:\sX \times \sX \rightarrow \sR$ be a positive semidefinite similarity function, with $k(\rvx, \rvx)=1$ for all $\rvx$, and let $\rmK \in \sR^{n\times n}$ denote a kernel matrix with entry $K_{i,j}=k(\rvx_i,\rvx_j)$. Denote by $\lambda_1, \cdots, \lambda_n$ the eigenvalues of $\rmK/n$. The Vendi Score ($\VS_k$) is defined as the exponential of the Shannon entropy of the eigenvalues of $\rmK/n$:
\begin{align}\label{eq:vendiscore}
\VS_k(\rvx_1,\cdots,\rvx_{n}) := \exp\left(-\sum_{i=1}^{n} \lambda_i \log\lambda_i\right),
\end{align}
following convention $0\log0=0$. The Vendi Score is computed from the eigenvalues of $\rmK/n$, instead of $\rmK$, so that its eigenvalues sum to one, and therefore the Shannon entropy is well-defined.

\section{VendiRL}

First, we adapt the Vendi Score to measuring skill diversity (Section~\cref{sec:effectivenumber}).
Then, we introduce our framework for learning diversely diverse skills (Section~\cref{sec:howvendirlworks}).
Lastly, we demonstrate that VendiRL can motivate different forms of diversity in skills according to a variable similarity function (Section~\cref{sec:whatskills}).

\subsection{A new way of thinking about skill diversity}
\label{sec:effectivenumber}

\textbf{Definition 2 (Effective number of unique skills).} We denote skills as specific configurations of policy parameters:
let $\rvtheta_i$ represent a particular skill corresponding to goal $i$, a specific configuration of the skills being learned, $\rvtheta$, from the set of all possible skills $\sTheta$.
A user-specified $k: \sTheta \times \sTheta \rightarrow \sR$ defines a similarity function according to which two skills are compared, such that $k(\rvtheta_i, \rvtheta_i)=1$ for all $i \in \left\{1,\cdots,n\right\}$, where $n$ represents the number of skills being learned, and $\rmK \in \sR^{n\times n}$ denotes a kernel matrix with entry $K_{i,j}=k(\rvtheta_i, \rvtheta_j)$.
Then, $\VS_{k^t}(\rvtheta_1,\cdots,\rvtheta_n)$ represents the diversity of an agent's skills at a particular point in time, $t$.
We call this quantity \emph{the effective number of unique skills}.

\begin{figure}[h!]
    \centering
    \begin{subfigure}{0.49\textwidth}
        \includegraphics[width=\textwidth]{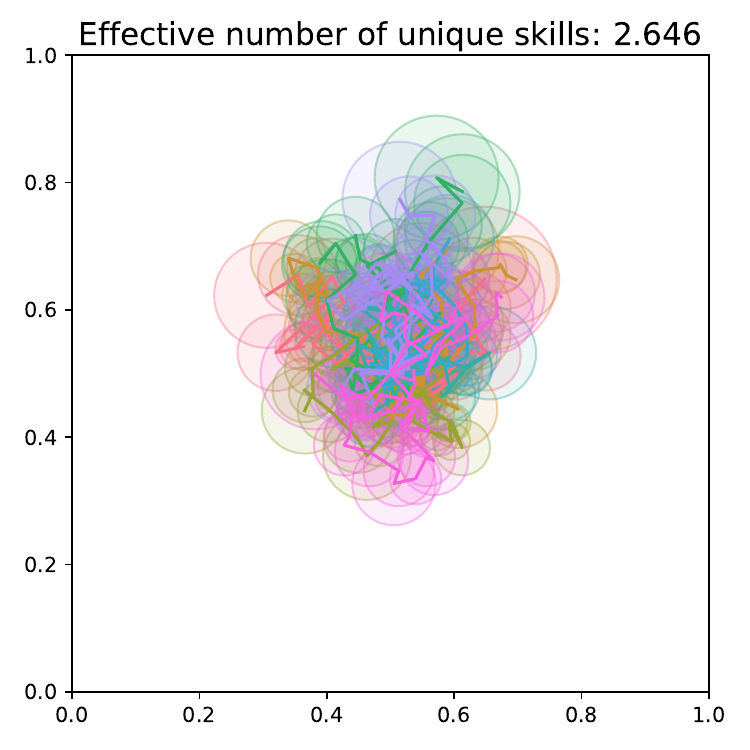}
        \caption{Random skills (no training).}
        \label{fig:randomskills}
    \end{subfigure}
    \hfill
    \begin{subfigure}{0.49\textwidth}
        \includegraphics[width=\textwidth]{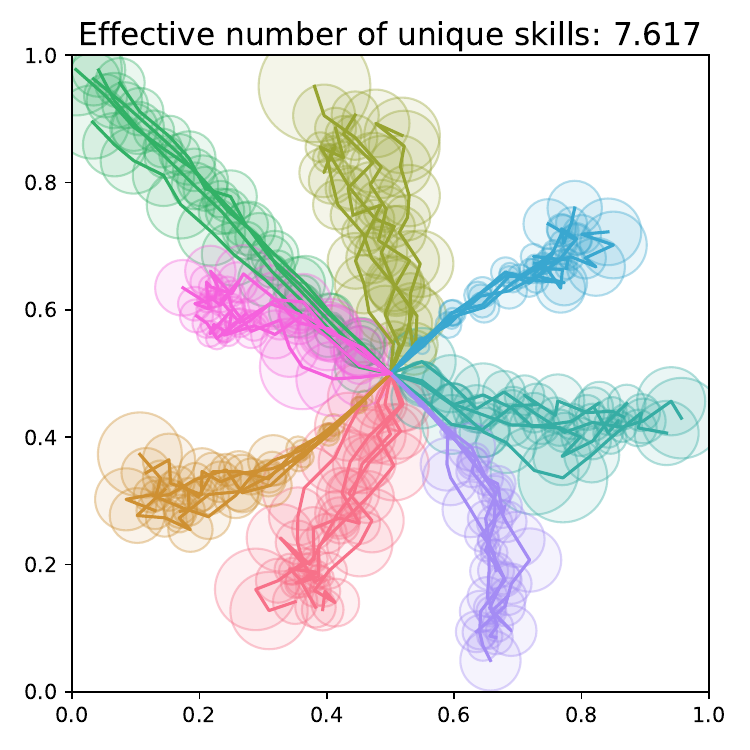}
        \caption{Skills trained with MISL (see Section~\cref{sec:misl}).}
        \label{fig:trainedskills}
    \end{subfigure}
    \caption{%
    Quantifying skill diversity with the Vendi Score.
    Here, the $F_1$ score is used as an example measure of similarity ($k$), estimating overlap between two skills in the feature space (see Appendix~\href{appendix:metricknnf1}{A}).
    Each of the eight skills is represented by five i.i.d. trajectories, and skills are differentiated by colour.}
    \label{fig:measuringdiversity}
\end{figure}

In our experiments, two skills are compared based on the trajectories of observations they induce.
Depending on the choice of $k$, we may use either full trajectories or summary statistics.
For instance, the mean of a trajectory over time captures information about the region of observation space the trajectory predominantly occupies, while the determinant of a trajectory's covariance matrix captures information about the trajectory's volume in the observation space.
In Figure~\cref{fig:measuringdiversity}, we used full trajectories, using a function $k$ that estimates \emph{overlap} of skills in the observation space.
This function estimates the $F_1$ score using a method originally intended for the finite approximation of data manifolds in the generative modelling literature \citep[]{kynkaanniemi2019improved}, which we adapted to \gls{rl} for approximating ``skill manifolds'' (see Appendix~\cref{appendix:metricknnf1}).
Other choices of $k$ are explored in Section~\cref{sec:whatskills}, where we provide examples of the sets of diverse skills they can motivate using VendiRL.

Figure~\cref{fig:measuringdiversity} illustrates the use of the Vendi Score in measuring the diversity of two distinct sets of skills.
Random skills (Figure~\cref{fig:randomskills}) result in extensive overlap when rolled out, so the effective number of unique skills is low (2.646).
In contrast, training skills with \acrlong{misl} (\acrshort{misl}; see Section~\cref{sec:misl}) results in skills that are discriminable in the observation space---clearly reflected in the Vendi Score (7.617), close to the maximum of eight (the agent has learned a set of eight skills).

Because the user can specify the similarity function used to compare skills, the Vendi Score lends itself to measuring any form of diversity that can be described mathematically.
This flexibility is also fundamental to our approach for motivating diversely diverse skills, which we elaborate on next.

\subsection{How VendiRL works}
\label{sec:howvendirlworks}

\begin{figure}[ht]
    \centering
    \begin{minipage}{0.39\textwidth}
        \centering
        \includegraphics[width=\textwidth]{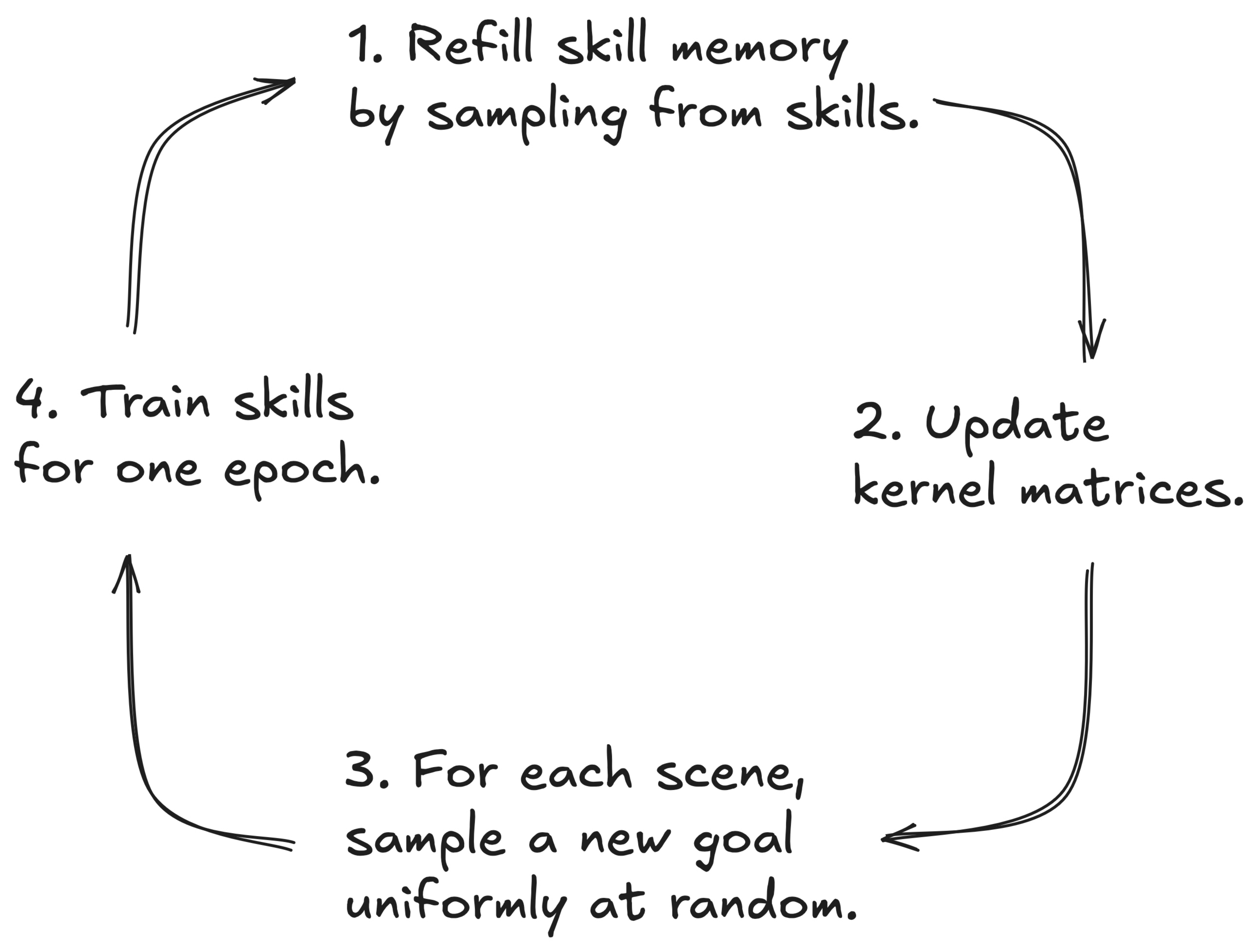}
    \end{minipage}%
    \hfill
    \begin{minipage}{0.59\textwidth}
        \begin{algorithm}[H]
            \small
            \DontPrintSemicolon
            \caption{VendiRL}
            \label{alg:vendirl}
            Define: similarity function $k$, number of skills $n$.\\
            Initialise: skills $\rvtheta$, skill memory $\rmM$, kernel matrix $\rmK$.\\
            \While{training}{
                Refill $\rmM$ by generating skill-trajectories from $\rvtheta$.\\
                Update $\rmK$ with $k$ (for all $n$ skills).\\
                Sample goal uniformly at random: $\rvg \sim \gU\{0,n-1\}$.\\
                \For{steps in epoch}{
                    Sample action: $\rva_t \sim \pi_\rvtheta(\rva_t \mid \rvs_t,\rvg)$.\\
                    Step environment: $\rvs_{t+1} \sim p(\rvs_{t+1} \mid \rvs_t,\rva_t)$. \\
                    Store observation in $\rmM$ (for current skill).\\
                    Update $\rmK$ with $k$ (for current skill).\\
                    Compute reward: $r_t := \exp \left(-\sum_{i=1}^{n} \lambda_i \log\lambda_i\right)$.\\
                    Use algorithm of choice to update $\rvtheta$.\\
                }
            }
        \end{algorithm}
    \end{minipage}
    \caption{%
    A schematic illustrating the high-level training loop (left) and our algorithm (right).
    }
    \label{fig:algorithm}
\end{figure}

The high-level training loop for an agent is as follows.
Following Definition 2 (Section~\cref{sec:effectivenumber}), we fix a similarity function $k$ and a number of skills to learn $n$.
Before any learning takes place, the agent generates a trajectory from each of its skills with the randomly initialised parameters $\rvtheta$ and stores the observations in a skill memory $\rmM$.
Following this, the agent computes pairwise similarities between its skills using $k$, which results in the kernel matrix $\rmK$.
Then, the agent selects a goal, $\rvg$, uniformly at random, following and updating the corresponding skill, $\rvtheta_\rvg$, for a fixed number of steps.

\begin{figure}[ht]
    \centering
    \begin{subfigure}{1.0\textwidth}
        \includegraphics[width=\textwidth]{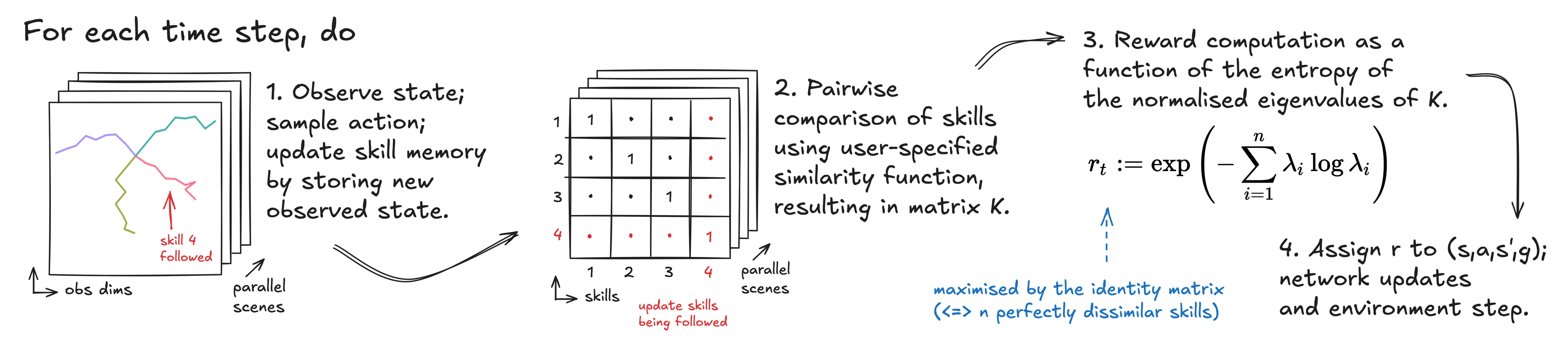}
    \end{subfigure}
    \caption{%
    A schematic illustrating a single \emph{VendiRL} training step within a parallelised training setup.
    }
    \label{fig:vendirlschematic}
\end{figure}

At each time step, the agent observes the current state, samples an action from its skill, and steps in the environment according to the environment's dynamics.
Observations are stored in the skill memory $\rmM$---which, in our experiments, is implemented as a queue with a maximum length equal to one episode.
This design allows the agent to track the most recent trajectory induced by each skill.
Consequently, the agent always replaces the observation at time $t$ from the previous episode with the observation at time $t$ from the current episode.
After storing the observation, the skill-similarity values in the kernel matrix $\rmK$ are updated for the elements corresponding to the skill currently being followed, $\rvtheta_\rvg$.
Specifically, the skill is compared pairwise with all other skills, $\left\{\rvtheta_1,\cdots,\rvtheta_n\right\}\backslash \left\{\rvtheta_\rvg\right\}$.
Following this update, the agent computes its goal-conditioned transition reward as defined below.

\textbf{Definition 3 (VendiRL reward).}
Given the kernel matrix, $\rmK \in \sR^{n\times n}$, consisting of pairwise similarity measurements of $n$ skills, $\rvtheta_1,\cdots,\rvtheta_n$, at time $t+1$, such that $K_{i,j}=k(\rvtheta_i, \rvtheta_j)$, the goal-conditioned transition reward is defined as the exponential of the Shannon entropy of the eigenvalues of $\rmK/n$ (following Equation~\cref{eq:vendiscore}, Section~\cref{sec:vendiscore}). That is, the Vendi Score, $\VS_k$, at time $t+1$:
\begin{align}\label{eq:vendireward}
r_t(\rvs_t, \rva_t, \rvs_{t+1}, \rvg_t) := \VS_{k^{t+1}}(\rvtheta_1,\cdots,\rvtheta_n).
\end{align}

Empirically, we found that transforming the reward worked well with some \gls{rl} algorithms.
Alternative formulations of the reward function, such as a \emph{time derivative} and \emph{penalty} are defined in Appendix~\cref{appendix:transformedrewards}.%

After computing the transition reward, skills ($\rvtheta$) are updated according to an optimiser of choice.
This motivates the agent to learn diverse skills according to $k$, thus driving a variable form of diversity.
After the skill has been followed for a fixed number of steps, the skill memory is refilled with new trajectories generated from $\rvtheta$,\footnote{Refilling the skill memory is a step taken to synchronise reward distributions across parallel scenes of an environment, a design choice discussed in Section~\cref{sec:scalingenv}.} the kernel matrix updated, a new goal selected, and the process repeated.
Our algorithm can be found in Figure~\cref{fig:algorithm}, and we illustrate a single training step in Figure~\cref{fig:vendirlschematic}.

\subsection{What skills are learned}
\label{sec:whatskills}

\begin{figure}[ht]
    \centering
    \begin{subfigure}{0.32\textwidth}
        \includegraphics[width=\textwidth]{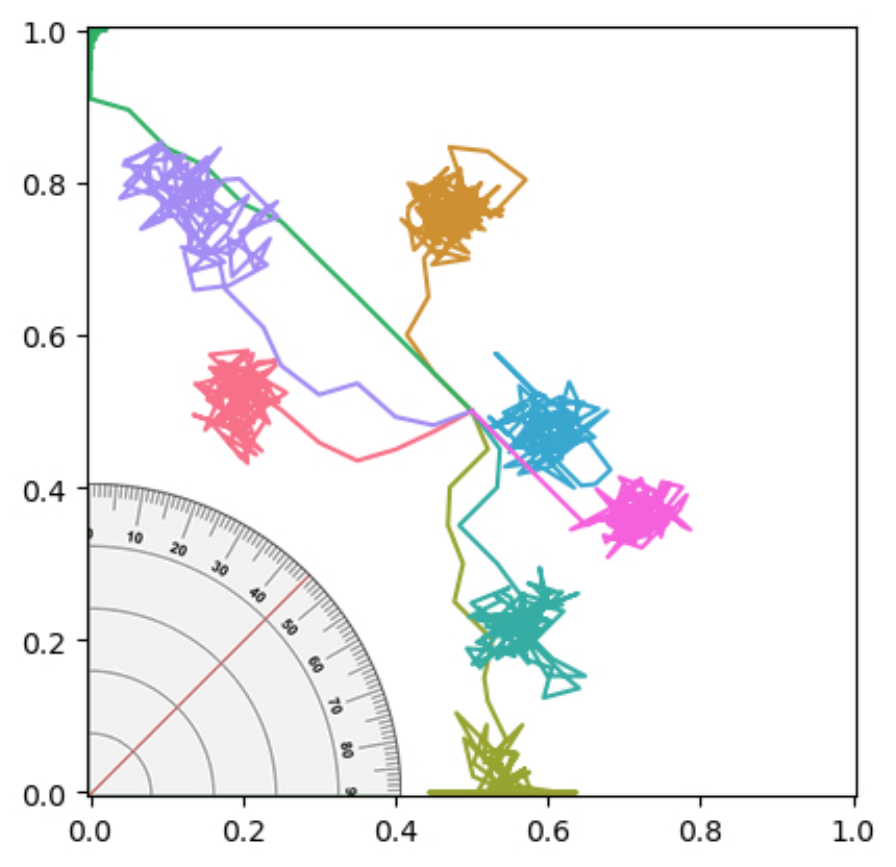}
    \end{subfigure}
    \hfill
    \begin{subfigure}{0.32\textwidth}
        \includegraphics[width=\textwidth]{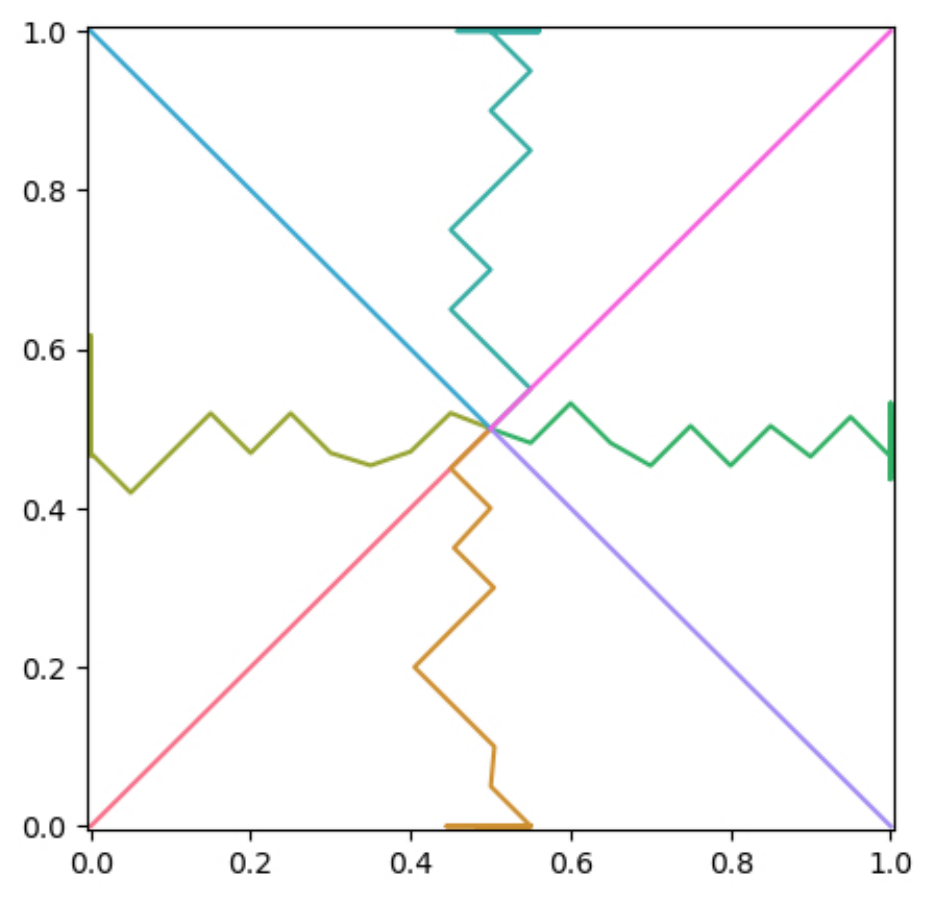}
    \end{subfigure}
    \hfill
    \begin{subfigure}{0.32\textwidth}
        \includegraphics[width=\textwidth]{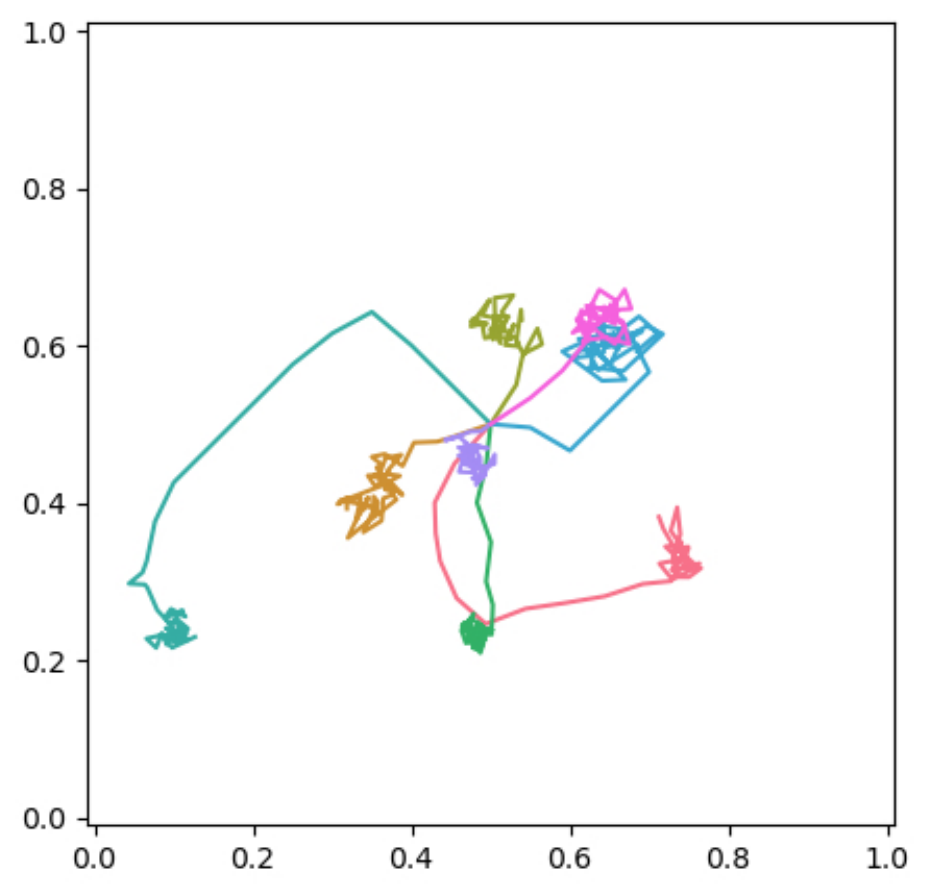}
    \end{subfigure}
    \caption{%
    Examples of \emph{VendiRL}-motivated skills in a $[0,1]^2$-bounded environment.
    In each case, an agent has learned eight skills that are differentiated by colour.
    Similarity between skills is measured as a function of (from left to right): cosine similarity, maximum mean discrepancy, covariance structure.
    }
    \label{fig:threesimilarityfunctions}
\end{figure}

To showcase the capabilities of VendiRL, we tested the framework using several distinct similarity functions.
As a proof of concept, we analysed the skills learned by agents in a simple 2D environment without extrinsic rewards, meaning that our agents were exclusively motivated by VendiRL rewards.

Figure~\cref{fig:threesimilarityfunctions} presents results from training three distinct sets of eight skills, each utilising a different similarity function.
The first function (left) is defined as cosine similarity between trajectory means, which drives the skills to angular separation around the origin.
The second function (middle) employs maximum mean discrepancy with a linear kernel, encouraging the skills to push away from one another.\footnote{This form of diversity is similar to that targeted by many state-of-the-art skill-diversity methods, visualised for comparison in Appendix~\cref{appendix:scalingexisting}, Figure~\cref{fig:existingdiversity}.}
The third function (right) measures similarity in covariance structure by computing absolute differences in the determinants of the skills' covariance matrices, %
thereby driving diversity in how widely the skills spread.
These similarity functions are described mathematically in Appendix~\cref{appendix:similarityfuncs}.

Beyond inducing distinct patterns of skill variation, Figure~\cref{fig:teasercombination} shows that VendiRL can target multiple notions of diversity simultaneously.
The 16 skills shown were trained using a similarity function that linearly combines cosine similarity and similarity in covariance structure.
This compositionality makes VendiRL a practical tool for designing composite diversity objectives and evaluation metrics.

\subsection{Challenges in scaling up}
\label{sec:scalingenv}

We follow \citet{freeman2021brax} by leveraging the auto-vectorisation capabilities of JAX \citep{bradbury2018jax} to parallelise data collection across a batch of independent environment scenes.
This changes the learning dynamics in several ways.
For instance, because goals are sampled independently per scene, an agent can pursue different skills in different scenes (and the same skill in multiple scenes).
Consequently, a single training epoch can expose the agent to a broad set of transitions spanning many skills, possibly including multiples of each skill that vary in state-action coverage.

Parallelisation massively accelerates the acquisition of diverse skills, but it introduces a challenge: different scenes induce different reward distributions.
To understand why this is the case and how we address it, we turn to the mechanisms underlying reward computation.

Each scene is allocated its own skill memory $\rmM$ and kernel matrix $\rmK$ rather than pooling them across scenes.
Pooling complicates credit assignment because: (1) the same skill can be active in multiple scenes simultaneously; (2) transition rewards depend on the current contents of the skill memory (i.e., the diversity among stored observations); and (3) skills are stochastic, so the same skill may visit different regions of the state space across scenes, especially early in training when the skills are near-random.
With a pooled memory, multiple observations induced by the same skill could be inserted into the memory at the same step, and the resulting reward would be broadcast to all those transitions, regardless of whether the underlying $(\rvs,\rva,\rvs',\cdot)$ are comparable.

Using independent per-scene memories resolves this credit-assignment issue but creates a new effect: a skill may be reinforced to increase diversity \emph{locally} in one scene while inadvertently encouraging similarity in another.
In other words, distinct memories induce distinct reward distributions.
To mitigate drift between reward distributions, the agent periodically synchronises memories: between epochs, each scene's memory is refilled with new trajectories generated from the shared skills $\rvtheta$.

More broadly, we view the design of $\rmM$ as central to scaling up the task of learning diverse skills.
Alternative parametrisations---for example, running averages of skills or learned skill representations against which new experience is compared---could amortise memory costs and thereby improve scaling.
Likewise, varying how skill information is shared across scenes provides a natural axis for experimentation, allowing us to study the trade-offs between reward stability and training throughput.

\subsection{Summary of contributions}
\label{sec:scalingdiversitypretraining}

Our work advances open-ended \gls{rl} in three main ways:
\begin{itemize}
    \item Unified, domain-agnostic evaluation: we adapt the Vendi Score to \gls{rl}, enabling comparable measures of skill diversity across algorithms, environments, and representations;
    \item Plug-in diversity objectives: a similarity function directly defines the reward, thus pursuing a different form of diversity does not require changing the system's architecture; and
    \item Pick-and-mix diversity: because the framework supports composable forms of diversity, it is a versatile tool for training and evaluating many forms of skill diversity.
\end{itemize}

\section{Related work}
\label{sec:relatedwork}

Recent progress in self‑supervised learning of diverse skills is heavily skewed towards a single methodological family; accordingly, we centre our review on that literature, then survey approaches to evaluating skill diversity, and close with related work outside the dominant paradigm.

\subsection{Mutual information skill learning}
\label{sec:misl}

\Gls{misl} is a well-studied subset of those intrinsically motivated skill acquisition methods focused on learning diverse sets of skills.
In \gls{misl}, an agent is tasked to maximise \gls{bmi}, that is, the mutual information between some representation of actions and some representation of states following those actions \citep[p.~3]{choi2021variational}.
Most often, this is done between skills (macro-actions) and some function of the agent's (skill-induced) trajectory, for example, the latest observation of the state.
To make the task tractable, most methods employ some form of \emph{variational} \gls{misl};
in this context, \gls{bmi} is approximated using a variational lower bound (e.g., \citealp[p.~2]{barber2003the}), and the problem of learning a diverse set of skills is formulated as a cooperative game between an actor and a learned discriminator model.\footnote{The formal details underlying variational \gls{misl} are given in Appendix~\cref{appendix:varmislformal}. For the interested reader, \citet{eysenbach2025rldm} provides an accessible tutorial on the topic.}

\citet[p.~3]{choi2021variational} provides an overview of variational approaches to \gls{misl}, also known as variational empowerment maximisation (for empowerment see \citealp{klyubin2005empowerment}).
Many different objectives for variational \gls{misl} have been proposed; examples of relevant empirical work include \citet{gregor2016variational}, \citet{warde-farley2019unsupervised}, \citet{eysenbach2019diversity}, \citet{hansen2020fast}, \citet{sharma2020dynamics-aware}, \citet{laskin2021urlb}, \citet{baumli2021relative}, \citet{yang2023behavior}, and \citet{zheng2025can}.
The works of \citet{eysenbach2022the} and \citet{reizinger2025skill} explain how \gls{misl} can be used to learn a ``universal'' representation for solving downstream tasks \citep[p.~5]{reizinger2025skill}.

One issue with scaling \gls{misl} is its reliance on a well-constructed feature space.\footnote{An extended discussion of the challenges \gls{misl} approaches face in scaling can be found in Appendix~\cref{appendix:scalingexisting}.}
Without it, agents may end up learning trivially diverse skills that neither result in observably distinct behaviours in the environment nor transfer effectively to downstream tasks---rendering the pretraining less worthwhile.
Although VendiRL does not fully resolve the issue, it can alleviate it by enabling the integration of additional prior knowledge.
Rather than partitioning the feature space into uniform subregions, VendiRL can target variable notions of diversity better suited to a downstream task domain.

\subsection{Evaluating skill diversity}

The dominant way to evaluate skill-diversity methods is via their utility: measure how well the skills learned during pretraining transfer to a selection of downstream tasks.
For example, some methods are evaluated by fine-tuning a set of learned skills on a set of downstream tasks (e.g., \citealp[p.~11]{gregor2016variational}; \citealp[pp.~6--7]{eysenbach2019diversity}; \citealp[pp.~6--7]{hansen2020fast}; \citealp[p.~4]{laskin2021urlb}; \citealp[pp.~7--8]{yang2023behavior}), and others using a hierarchical controller to control the repertoire of learned skills on a set of downstream tasks (e.g., \citealp[pp.~7--8]{eysenbach2019diversity}, \citealp[pp.~9--10]{sharma2020dynamics-aware}; \citealp[p.~6737]{baumli2021relative}; \citealp[p.~8]{shafiullah2022one}; \citealp[pp.~8--10]{park2024metra}).

Beyond utility, other common approaches for evaluation include qualitative analyses visualising skill trajectories (e.g., \citealp[p.~5]{eysenbach2019diversity}; \citealp[pp.~6--8]{sharma2020dynamics-aware}; \citealp[Appendix A, pp.~19--23]{gu2021braxlines}; \citealp[pp.~7--8]{park2024metra}), measuring state space coverage (e.g., \citealp[p.~8]{zheng2023cim}; \citealp[pp.~8--9]{park2024metra}), and scoring on an agent on the diversity reward used for training (e.g., \citealp[p.~8]{gregor2016variational}, \citealp[Appendix D.1, p.~15]{eysenbach2019diversity}; \citealp[pp.~6--8]{choi2021variational}).

Other proposed proxies include: the effective number of skills (\citealp[p.~6; Appendix D, pp.~17--18]{eysenbach2019diversity}), measured as the exponential of the Shannon entropy of \emph{the goal-selection policy}, resulting in a quantity representing the effective number of skills \emph{being considered for selection by an agent at a given time}; \acrlong{lgr} (e.g., \citealp[pp.~6--8]{choi2021variational}; \citealp[Appendix B, p.~25]{gu2021braxlines}); distributions over task reward \citep[Appendix D.3, p.~16]{eysenbach2019diversity}; skill-separability metrics (e.g., \citealp[pp.~5,~7]{yang2024task}); per-dimension goal achievement \citep[pp.~7--10]{warde-farley2019unsupervised}; Diffusion Time (e.g., \citealp[Appendix B, p.~13]{machado2017a}); \acrlong{snd} \citep{bettini2024system}; and \gls{qd} scores (see \citealp[p.~9]{pugh2016quality}).

While each measure usefully captures some facet of skill diversity, most make a hard commitment to a single notion and thus do not support evaluating variable forms of diversity.
With the exception of utility-based and \acrshort{qd}-like measures, they rarely enable the user to specify what ``diversity'' should mean for a particular domain.
Several evaluation metrics can, in principle, reflect different forms of diversity by changing the feature representation, but in practice the representation is constrained by the domain and determined by what works well for learning the task.
In contrast, our use of the Vendi Score brings a unified, domain‑agnostic notion of skill diversity to \gls{rl}, enabling consistent evaluation of behavioural diversity under a variable similarity function specified by the user.

\subsection{Other approaches for learning diverse skills}

\paragraph{Regularity-based rewards} 
\citet{sancaktar2023regularity} use regularity as an intrinsic reward.
In their approach, the state space is factorised into objects, mapped to a multiset of symbols (e.g., discretised positions, colours).
Varying this mapping induces distinct types of structured behaviour by changing what is considered ``regular'', though the aim is not explicitly to support variable notions of diversity.

\paragraph{Automatic curricula}  
\citet{openai2021asymmetric} use asymmetric self-play in a goal-conditioned setting: one agent aims to propose challenging tasks, the other aims to solves them.
This interaction results in a curriculum that uncovers complex tasks and the diverse skills to solve them.
\citet{colas2019curious} partition the sensory space into subspaces (``modules'') and design a system for agents to self-organise curricula over modules, prioritising subspaces of goals they are increasingly or decreasingly successful in reaching.
Thereby, agents can learn the skills needed to tackle a diverse range of goals.

\paragraph{Incremental expansion of capacities} 
\citet{shafiullah2022one} grow a diverse skill repertoire sequentially (each its own policy), optimising new skills for high state entropy w.r.t. existing skills and low entropy within-skill, promoting diversity in skills while preserving controllability.
\citet{pong2020skew-fit} learn a maximum-entropy goal distribution via importance-weighted training of a goal generator such that rarely visited states are given more weight.
This can result in uniform coverage of states, inducing a diverse range of skills when reaching diverse goals requires diverse behaviours.

\paragraph{Laplacian-based approaches}
Eigenoptions \citep[]{machado2017a} are skills derived from  temporal properties of the state space though the eigendecomposition of a matrix representation of its underlying graph.
The decomposition yields intrinsic reward functions for learning skills that operate at different time scales.
\citet{chen2023a} unify Laplacian-based methods with \gls{misl}, extracting the benefits of the two into a powerful framework for learning skills that are diverse and achieve high state coverage.

\paragraph{Language-guided discovery} 
\citet{rho2025language} use language models to guide skill discovery, using \emph{language-distance} as a proxy for semantic distance to learn ``semantically diverse'' skills.

\paragraph{Broader context} 
The landscape of self-supervised \gls{rl} for diverse skills is rapidly evolving; \citet[Appendix A, pp.~17--18]{park2024metra} provide an extensive list of recent approaches.

\section{Limitations and future work}
\label{sec:limitationsfuturework}

\paragraph{Downstream utility} 
We have not evaluated whether the learned skills are useful on future tasks.
It remains an open question how skill transfer depends on the choice or mixture of similarity functions.

\paragraph{Scalability and stability} 
Regarding scalability, computing the eigenvalues of a $\sR^{n\times n}$ kernel matrix has a time complexity of $\gO(n^3)$; scaling to large sets of skills may require approximation.
Regarding stability, in the case of many similarity functions, the objective admits multiple behaviourally distinct optima, making it hard to exactly predict where the skills converge under different initialisations.

\paragraph{Future directions}
First, evaluate transfer learning: how useful are VendiRL-motivated skills on downstream tasks? Can properties of downstream tasks be mapped to effective similarity functions or their mixtures?
Second, meta-learning: can we rely less on human priors by training learners to select and weight different forms of diversity autonomously?
Third, analyse key design choices underlying the framework: choice of skill representation and skill memory, mechanisms to mitigate drift between reward distributions under parallelism, and the \gls{rl} algorithm used for training the goal-conditioned policy.
Lastly, we aim to deepen our understanding of how different similarity functions capture different notions of behavioural diversity.
This knowledge, in turn, can support a widely acknowledged need: evaluation and fair comparison of the capabilities of open-ended systems.

\section{Conclusion}
\label{sec:conclusion}

We introduced VendiRL, a self-supervised \gls{rl} framework that centres both skill learning and evaluation on a user-specified similarity function and the Vendi Score.
Our proof-of-concept shows that different similarity functions induce distinct behaviours, decoupling what counts as diverse from how rewards are computed.
Adapting the Vendi Score to \gls{rl} further supports evaluation and benchmarking: the same metric can quantify outcomes across algorithms, representations, and domains, assuming a comparable number of skills.
This opens up new uses, for example: direct comparisons of existing skill-learning methods under shared notions of diversity, use of diversity as a reward signal for \acrlong{ued} (e.g., generating variants based on the skill diversity they afford), and construction of skill-diversity benchmarks with variable and composable diversity targets.
In short, VendiRL turns what is considered diverse into a controllable and optimisable design choice.

\begin{ack}
I thank the \href{https://www.aalto.fi/en/department-of-computer-science/autotelic-interaction-research}{Autotelic Interaction Research (AIR)} group at Aalto University,
especially Nadia Ady, Perttu Hämäläinen, and Christian Guckelsberger.
I am particularly grateful to Christian for introducing me to the Vendi Score and for feedback on parts of the draft.
I extend my gratitude to Luigi Acerbi and Antti Honkela for their \href{https://www.cs.helsinki.fi/u/ahonkela/teaching/compstats1/book/index.html}{computational statistics} course at the University of Helsinki, where I learned some neat numerical linear algebra tricks applied in this work;
the reviewers for their insightful comments; 
and my three-month old, for kindly agreeing to sleep in my lap while I wrote.
This work was funded in part by the Research Council of Finland's NEXT-IM project (grant no. 349036).
\end{ack}

{
\bibliographystyle{plainnat}
\bibliography{references}
}

\newpage
\appendix

\section{Skill diversity as minimal overlap in the feature space}
\label{appendix:metricknnf1}

As one example of measuring skill diversity with the Vendi Score, we employ a skill-similarity measure derived from a $k$-nearest neighbors formulation of \emph{precision} and \emph{recall} developed by \citet{kynkaanniemi2019improved}.
Note that $k$ here denotes a scalar variable, deviating from the use of $k$ in the main body of the paper to represent a kernel function.

To compute the overlap between two stochastic skills, $\rvtheta_a$ and $\rvtheta_b$, we collect two samples of $N$ trajectories, $\rmX_a \sim \rvtheta_a$ and $\rmX_b \sim \rvtheta_b$, respectively. In our case, each trajectory is a $\sR^{T \times D}$ matrix, where $T$ represents the length of the trajectory in time steps and $D$ the number of observation dimensions in the given environment. For each skill, the samples are concatenated, so we end up with an $\sR^{NT \times D}$ matrix consisting of $NT$ observation vectors each denoted $\rvx$. Then, for each set of observation vectors, $\rmX \in \{\rmX_a,\rmX_b\}$, we compute pairwise Euclidean distances between all observations in the set and, for each observation vector, form a hypersphere with radius equal to the distance to its $k$th nearest neighbour. Together, these hyperspheres form an estimate of the true skill manifold in the observation space (illustrated in Figure~\cref{fig:measuringdiversity}). Following \citet[p.~3]{kynkaanniemi2019improved}, we use a binary function to determine whether a given observation is located within this manifold:
\begin{align}\label{eq:knnf1binary}
f(\rvx, \rmX) :=
    \begin{cases}
        1, & \text{if } || \rvx-\rvx' || \leq || \rvx'-\mathrm{NN}_k(\rvx',\rmX) || \text{ for at least one } \rvx' \in \rmX \\
        0, & \text{otherwise,}
    \end{cases}
\end{align}
where $\mathrm{NN}_k(\rvx',\rmX)$ represents the $k$th nearest observation from $\rvx' \in \rmX$. Then, as in \citet[p.~3]{kynkaanniemi2019improved}, precision and recall are defined respectively as:
\begin{align}\label{eq:knnf1precisionrecall}
\mathrm{pr}(\rmX_a, \rmX_b) := \frac{1}{|\rmX_b|} \sum_{\rvx_b \in \rmX_b} f(\rvx_b, \rmX_a),
\quad
\mathrm{re}(\rmX_a, \rmX_b) := \frac{1}{|\rmX_a|} \sum_{\rvx_a \in \rmX_a} f(\rvx_a, \rmX_b).
\end{align}

Then, the overlap between $\rmX_a$ and $\rmX_b$ is given by the harmonic mean of precision and recall:
\begin{align}\label{eq:knnf1f1}
F_1(\rmX_a, \rmX_b) := \frac{2 \times \mathrm{pr}(\rmX_a, \rmX_b) \times \mathrm{re}(\rmX_a, \rmX_b)}{\mathrm{pr}(\rmX_a, \rmX_b)+\mathrm{re}(\rmX_a, \rmX_b)},
\end{align}
such that $F_1=1 \iff \rmX_a=\rmX_b$, and $F_1 \to 0$ indicates little to no overlap.

Precision measures the proportion of observations drawn from skill $\rvtheta_b$ that fall within the estimated support of $\rvtheta_a$, and recall measures the proportion of observations drawn from skill $\rvtheta_a$ that fall within the estimated support of $\rvtheta_b$. The $F_1$ score weights the two measures equally. After computing pairwise scores between skills, their overall diversity---or, the effective number of unique skills---is given by computing the Vendi Score from the kernel matrix (as detailed in Section~\cref{sec:effectivenumber}).

\section{Transforming VendiRL rewards}
\label{appendix:transformedrewards}

\subsection{Time derivative}

To reward an agent for \emph{increasing} the diversity of its skills between two consecutive time steps, the VendiRL rewards can be defined as $r_t(\rvs_t, \rva_t, \rvs_{t+1}, \rvg_t) := \VS_{k^{t+1}}(\rvtheta_1,\cdots,\rvtheta_n)-\VS_{k^{t}}(\rvtheta_1,\cdots,\rvtheta_n)$.

\subsection{Penalty}

The Vendi Score lies in $[1,n]$.
Thus, a VendiRL reward can be easily transformed into a \emph{penalty} with $r_t(\rvs_t, \rva_t, \rvs_{t+1}, \rvg_t) := \VS_{k^{t+1}}(\rvtheta_1,\cdots,\rvtheta_n)-n$.
We also found $(f\circ g)(r_t)$, where $f(x):=\log(x)$ and $g(x):=x/n$, such that $g:[1,n]\rightarrow[1/n,1]$ and $f:[1/n,1]\rightarrow[\log(1/n), 0]$, to work well.

\section{Similarity functions included in this paper}
\label{appendix:similarityfuncs}

As before, we denote skills as specific configurations of policy parameters: let $\rvtheta_i$ represent a particular skill corresponding to goal $i$, a specific configuration of the skills being learned, $\rvtheta$, from the set of all possible skills $\sTheta$.
Then, $k(\rvtheta_a,\rvtheta_b)$ measures the similarity between $\rvtheta_a$ and $\rvtheta_b$ such that $k: \sTheta \times \sTheta \rightarrow \sR$.

Notably, in case one wishes to enforce $k(\rvtheta_i, \rvtheta_i)=1$ for all $i \in \left\{1,\cdots,n\right\}$, where $n$ denotes the number of skills being learned, outputs from some choices of $k$ will have to be scaled accordingly.

\subsection{Cosine similarity}

In addition to the above, let $\rvmu_i$ represent the mean of a trajectory, over time, induced by skill $\rvtheta_i$.
Then, the cosine similarity between skills $\rvtheta_a$ and $\rvtheta_b$ is given by
\begin{align}\label{eq:cossim}
k(\rvtheta_a,\rvtheta_b) := \frac{\rvmu_a\cdot\rvmu_b}{\left\Vert\rvmu_a\right\Vert\left\Vert\rvmu_b\right\Vert}.
\end{align}

In an unbounded feature space, the output of $k$ is in $[-1,1]$.
Left unscaled, $k(\rvtheta_a,\rvtheta_b)=1$ when the two means have no angular separation and $k(\rvtheta_a,\rvtheta_b)=-1$ when opposite one another.

\subsection{Maximum mean discrepancy}

Again, where $\rvmu_i$ denotes the mean of a trajectory induced by $\rvtheta_i$, the maximum mean discrepancy (with linear kernel) between skills $\rvtheta_a$ and $\rvtheta_b$ is given by
\begin{multline}\label{eq:mmd}
k(\rvtheta_a,\rvtheta_b) := (f\circ g)(\rvmu_a,\rvmu_b), \quad \text{where} \\ f(x) := \frac{1}{\exp(x)} \quad \text{and} \quad g(\rvmu_a,\rvmu_b) := \left\Vert\rvmu_a-\rvmu_b\right\Vert.
\end{multline}

The function $g$ is derived as follows \citep[]{danica2017mmd}.
\Gls{mmd} is defined based on a feature map $\varphi: \sX \rightarrow \sH$, where $\sH$ is some Hilbert space.
Then, when $\sX=\sH=\sR^d$ and $\varphi(x)=x$, corresponding to a linear kernel, we have
\begin{align*}
\text{MMD}(P,Q) &= \left\Vert \E_{X\sim P}\left[\varphi(X)\right] - \E_{Y\sim Q}\left[\varphi(Y)\right] \right\Vert_\sH \\
&= \left\Vert \E_{X\sim P}\left[X\right] - \E_{Y\sim Q}\left[Y\right] \right\Vert_{\sR^{d}} \\
&= \left\Vert \mu_P - \mu_Q \right\Vert_{\sR^{d}}.
\end{align*}

Computing the distance ($g$) and converting the output into a similarity $(f)$ results in $k(\rvtheta_a, \rvtheta_b)=1$ when the two means are equal and a quantity in $(0,1)$ otherwise (tending to zero as the distance grows in magnitude).

\subsection{Covariance structure}

Given a trajectory of observations induced by skill $\rvtheta_i$, we denote by $\rmSigma_i$ the corresponding sample covariance matrix.
Then, the similarity in covariance structure is defined as
\begin{multline}\label{eq:detcovmat}
k(\rvtheta_a,\rvtheta_b) := (f\circ g)(\rmSigma_a,\rmSigma_b), \quad \text{where} \\ f(x) := \frac{1}{\exp(x)} \quad \text{and} \quad g(\rmSigma_a,\rmSigma_b) := \left|\det(\rmSigma_a)-\det(\rmSigma_b)\right|.
\end{multline}

The determinant $\det(\rmSigma_i)$ is evaluated using Cholesky decomposition as follows \citep[]{honkela2020computational}.
The symmetric positive definite matrix $\rmSigma_i$ can be represented as $\rmSigma_i=\rmL\rmL^T$, where $\rmL$ is a lower-triangular matrix.
Then, using basic properties of the determinant and the logarithm, we obtain
\begin{align*}
\log\det \rmSigma_i &= \log\left(\det(\rmL\rmL^T)\right) = \log\left(\det\rmL\det\rmL^T\right) \\
&= \log\left((\det\rmL)^2\right) = 2\log(\det\rmL) \\
&= 2\log\left(\prod_{i=1}^d l_{ii}\right) = 2\sum_{i=1}^d\log(l_{ii}).
\end{align*}

The determinant of the skill's covariance matrix, as the product of its eigenvalues---each of which represents the magnitude of the skill's spread on a principal axis---captures information about the skill's \emph{volume} in feature space.
Taking the absolute difference ($g$) and converting the output into a similarity $(f)$ results in $k(\rvtheta_a, \rvtheta_b)=1$ when the two determinants are equal and a quantity in $(0,1)$ otherwise (tending to zero as the absolute difference grows in magnitude).

\section{Extended discussion of MISL}

\subsection{Variational MISL}
\label{appendix:varmislformal}

Formally, variational \gls{misl} approximates the lower bound on mutual information using a discriminator $q$ with parameters $\rvphi$, between the goal-defining variable, $\rvg$, 
and some function of the trajectory induced by the corresponding skill.
For concreteness, we represent the output of this function by a common choice, the observed successor state, $\rvs'$, determined by the state-transition distribution, $p_\rvtheta(\rvs'\mid \rvs,\rvg)$, conditional on the skill-defining parameters $(\rvtheta, \rvg)$.
Then, the objective is to maximise
\begin{align}
\label{eq:varmislobjective}
\gF(\rvtheta,\rvphi) := \E_{\rvg\sim p(\rvg),\rvs'\sim p_\rvtheta(\rvs'\mid \rvs,\rvg)}[\underbrace{\log q_\rvphi(\rvg\mid \rvs')}_\text{($\alpha$)} \underbrace{-\log p(\rvg)}_\text{($\beta$)}].
\end{align}
\begin{itemize}
  \item[($\alpha$)] Given the agent's ($\rvtheta$) behaviour, the discriminator ($\rvphi$) tries to predict which skill the agent is following. The agent is rewarded for learning predictable and thus diverse skills: successfully discriminating skills requires the agent to observe distinct regions of the feature space. %
  \item[($\beta$)] This term is maximised in expectation when skills are selected uniformly at random. If the agent does not learn $p(\rvg)$, it is a common choice to fix it to a uniform distribution, resulting in a constant term. For more detail on fixing versus learning $p(\rvg)$, see \citet{lintunen2024diversity}.
\end{itemize}

\subsection{Problems in scaling MISL}
\label{appendix:scalingexisting}

\begin{figure}[ht]\captionsetup[subfigure]{font=scriptsize}
    \centering
    \begin{subfigure}{0.23\textwidth}
        \includegraphics[width=\textwidth]{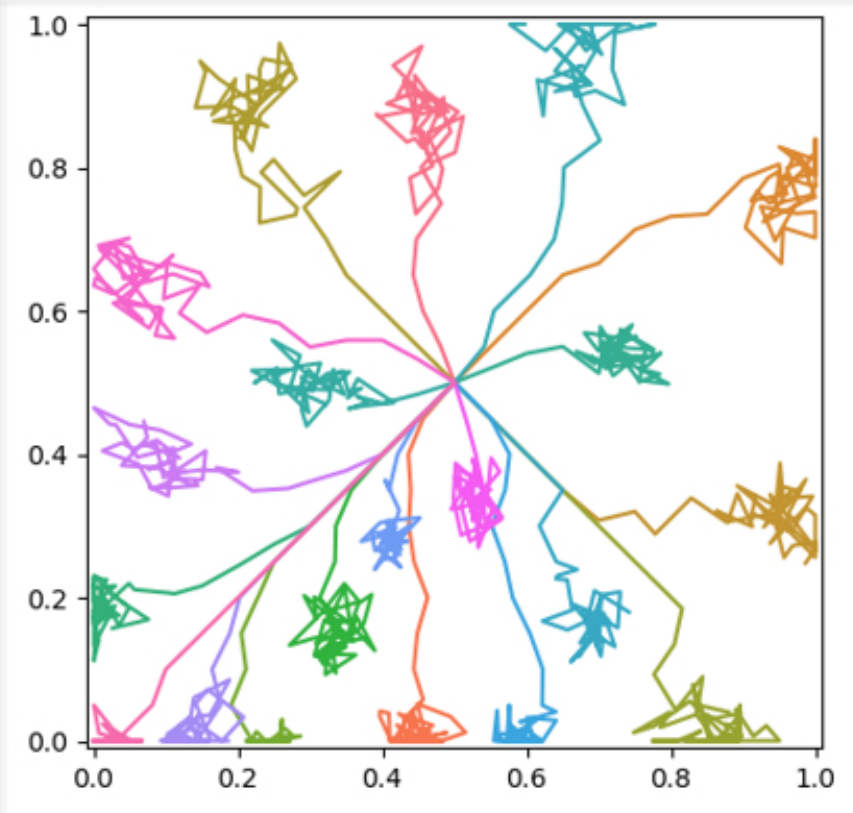}
        \caption{\emph{Diversity is All You Need} \citep{eysenbach2019diversity}, a prototypical \acrshort{misl} approach, as part of which an agent maximises the mutual information between skills and the states they induce.}
        \label{fig:diayn}
    \end{subfigure}
    \hfill
    \begin{subfigure}{0.23\textwidth}
        \includegraphics[width=\textwidth]{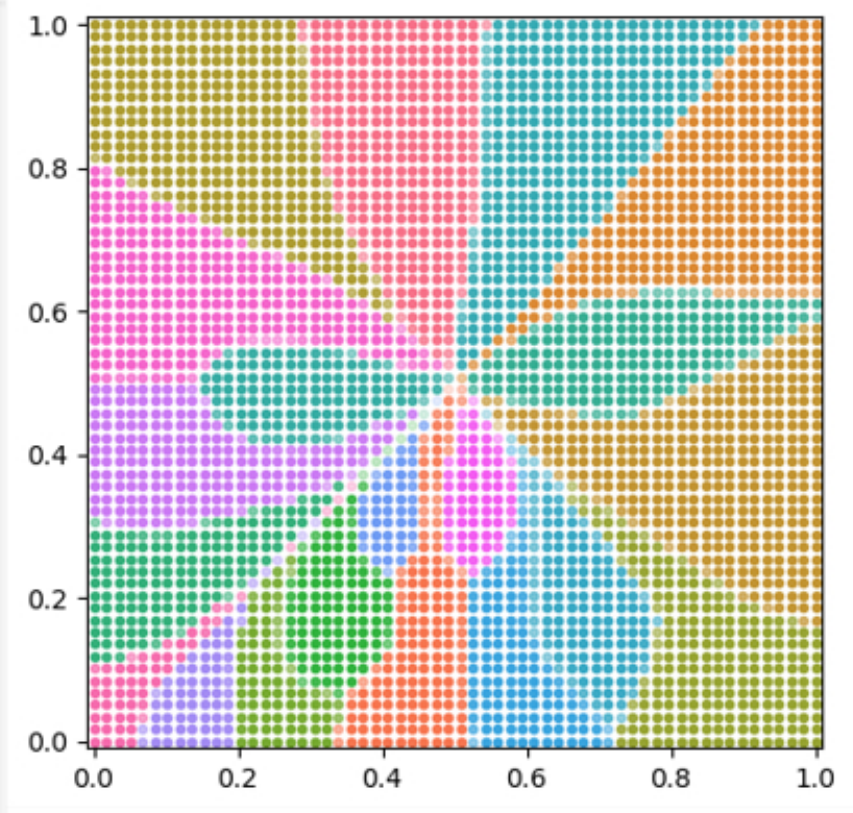}
        \caption{The corresponding decision boundary of the learned classifier used to predict skills from observations. To maximise rewards, the agent has has to be correct in and certain of its predictions.}
        \label{fig:diayndisc}
    \end{subfigure}
    \hfill
    \begin{subfigure}{0.23\textwidth}
        \includegraphics[width=\textwidth]{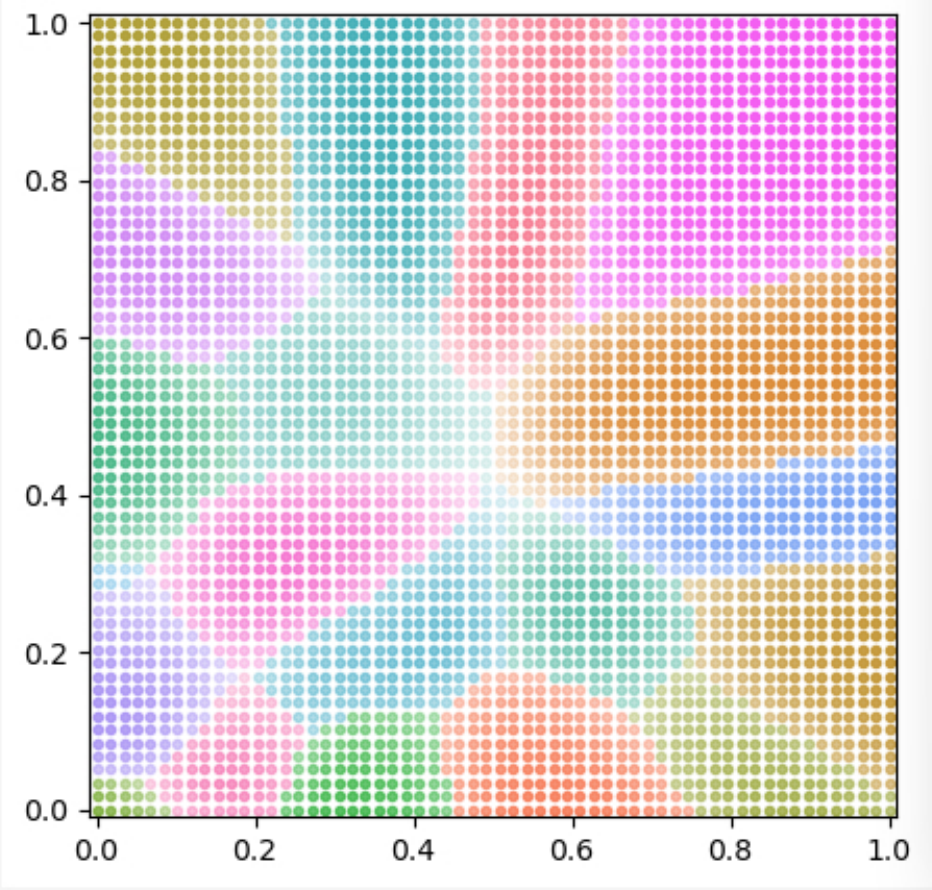}
        \caption{Spectral normalisation can be used to enforce smoothness in the decision landscape \citep[pp.~5--6]{choi2021variational}. This makes the task harder to solve, but its use helps avoid overfitting to noise.}
        \label{fig:diaynsndisc}
    \end{subfigure}
    \hfill
    \begin{subfigure}{0.23\textwidth}
        \includegraphics[width=\textwidth]{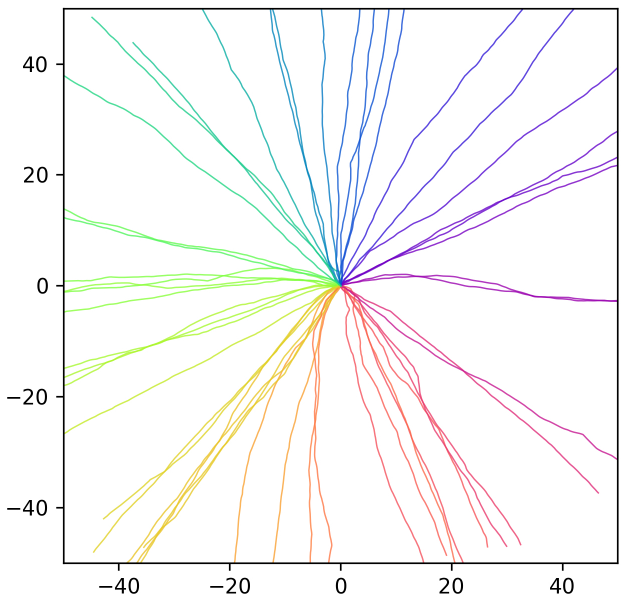}
        \caption{\emph{Contrastive Successor Features} (\citealp[visualisation from cited paper with no changes]{zheng2025can}) maximises the mutual information between transitions and skills, fixing dithering issues.}
        \label{fig:csf}
    \end{subfigure}
    \caption{Evolution of mutual information skill learning. Some problems and subsequent solutions.}
    \label{fig:existingdiversity}
\end{figure}

\subsubsection{Having to determine the number of skills}

Learning \emph{the} optimal initialisation for the set of all unknown reward functions in an environment would typically require an agent to learn a large number of skills, making the pretraining computationally demanding if not intractable (to do optimally).\footnote{\citet[p.~8]{reizinger2025skill} show that the canonical way of fixing a small number of skills and drawing skills uniformly at random during training is insufficient for learning the ground-truth features of an environment.}
Instead, it is common to try to approximate the optimal initialisation by fixing some small number of skills, $n$, and learning $n$ skills from the set of all possible skills.
However, not all methods motivate agents to go beyond what is required to discriminate the skills.
If $n$ is too small, the skills can start dithering in discriminable regions (e.g., Figure~\cref{fig:diayn}).
Further, if the dimensionality of the feature space is high, finding discriminable regions in that space can become trivial---without increasing the number of skills---due to its increase in volume.
Discriminability-motivated learning can thus be challenging to implement effectively in complex environments.
That said, some \gls{misl} methods have been developed recently to address this problem (e.g., Figure~\cref{fig:csf}), but they still rely on the user's intuition of how many skills an agent should learn.

\subsubsection{The discriminator can overfit to random skills}

Since the skills are randomly initialised, and the discriminator is typically a neural network with high expressive power, the neural network can easily overfit to the random skills.
This can lead to a non-smooth decision landscape of the discriminator (e.g., Figure~\cref{fig:diayndisc}), which can result in the skills converging to a suboptimal solution with respect to their diversity.
One proposed solution is to use spectral normalisation for regularising the discriminator (\citealp[pp.~5--6]{choi2021variational}, referring to the work of \citealp{miyato2018spectral}).
This enforces smoothness in the discriminator's decision landscape (e.g., Figure~\cref{fig:diaynsndisc}); making the task harder to solve, but helping to prevent overfitting to near-random skills.
While spectral normalisation clearly offers a solution to the problem of overfitting to random skills, we lack research on the trade-offs between different levels of discriminator expressivity (learning to discriminate skills effectively) and regularisation (preventing overfitting).

\subsubsection{Relying on the assumption of a well-structured feature space}

The use of \gls{misl} often relies on the user's supervision for structuring the feature space.
In practice, this means that when implementing a \gls{misl} agent, some dimensions of the feature space are ignored by design.
With some specific subset of downstream reward functions in mind, the user determines the most appropriate subset using their prior knowledge of task-relevant features.
In such cases, agents maximise diversity in some low-dimensional subset of the space of behaviours (e.g., \citealp[p.~7]{eysenbach2019diversity}).
Without such supervision, agents can end up learning trivially diverse skills that neither translate to observably diverse behaviours in the environment, nor transfer to downstream tasks effectively enough to make the pretraining worthwhile.
While automated solutions to this problem have been proposed, such as learning a feature representation that supports transfer learning (e.g., \citealp{smith2022learning}), they still rely on external supervision to the extent of specifying the set---or at the very least, the distribution---of future tasks.

\end{document}